\documentclass[
]{ceurart}

\usepackage{subcaption}

\usepackage{listings}
\begin{document}

\copyrightyear{2025}

\copyrightclause{Copyright for this paper by its authors. Use permitted under Creative Commons License Attribution 4.0 International (CC BY 4.0).}

\conference{Preprint. Submitted to arXiv.}


\title{An interpretable prototype parts-based neural network for medical tabular data}


\author[1]{Jacek Karolczak}[%
orcid=0000-0001-5414-960X,
email=jacek.karolczak@cs.put.poznan.pl
]
\address[1]{Poznan University of Technology, Institute of Computing Science, ul. Piotrowo 2, 60-695 Poznań, Poland}

\author[1]{Jerzy Stefanowski}[%
orcid=0000-0002-4949-8271,
email=jerzy.stefanowski@cs.put.poznan.pl
]

\cortext[1]{Corresponding author.}

\begin{abstract}
    The ability to interpret machine learning model decisions is critical in such domains as healthcare, where trust in model predictions is as important as their accuracy. Inspired by the development of prototype parts-based deep neural networks in computer vision, we propose a new model for tabular data, specifically tailored to medical records, that requires discretization of diagnostic result norms. Unlike the original vision models that rely on the spatial structure, our method employs trainable patching over features describing a patient, to learn meaningful prototypical parts from structured data. These parts are represented as binary or discretized feature subsets. This allows the model to express prototypes in human-readable terms, enabling alignment with clinical language and case-based reasoning. Our proposed neural network is inherently interpretable and offers interpretable concept-based predictions by comparing the patient's description to learned prototypes in the latent space of the network. In experiments, we demonstrate that the model achieves classification performance competitive to widely used baseline models on medical benchmark datasets, while also offering transparency, bridging the gap between predictive performance and interpretability in clinical decision support.
\end{abstract}



\maketitle

\section{Introduction}
\label{sec:introduction}

Machine learning (ML) has been increasingly used in medicine for many decades, in particular to improve diagnostic accuracy, predict patient outcomes, and support clinical decision making by uncovering complex patterns in medical data \cite{christodoulou2019logisticregression}. Early applications of machine learning prioritized inherently interpretable models that provided symbolic knowledge representations, such as \textit{decision trees} (DT) and rule-based systems \cite{podgorelec2002decisiontree}. Encouraged by the initial successes of these approaches, researchers began addressing more complex problems using more advanced models such as \textit{random forests} (RF), other ensembles, or even hybrid approaches \cite{breiman2001rf}. Although these ML systems offer an improvement in predictive performance \cite{vlachas2022rfmedicine}, they do so at the expense of transparency and interpretability \cite{bharati2024healthcarereview}.

Nowadays, in many tasks, \textit{deep neural networks} (NN) have become the most popular approach, particularly for analyzing modalities such as images, time series, or text data. However, a significant portion of clinical work still relies on tabular data, where the application of deep learning models, due to their black-box nature, is less widespread and less appreciated. In the healthcare domain, the reluctance to adopt NN is partially driven by the difficulty in interpreting their decision-making processes, making it challenging for physicians to analyze, validate, and ultimately trust their results in real-world applications. As a result, there has been growing interest in using \textit{Explainable AI} (XAI) techniques~\cite{bodria2023xai} to make machine learning models more transparent and understandable for clinical use.

Currently, the landscape of XAI is dominated by feature importance methods such as SHAP \cite{lundberg2017shap} and LIME \cite{ribeiro2016lime} being among the most widely used. However, despite their popularity, these approaches often produce abstract and incomprehensible explanations, even for machine learning experts, and can be particularly challenging for physicians to understand~\cite{longo2024manifesto}. As a result, there is growing interest in alternative paradigms that provide more intuitive and human-understandable insights aligned with the way physicians reason about the patient and the diagnosis.

In this context, \textit{prototypes} -- instances that represent groups of similar examples -- have emerged as a particularly promising explanation technique \citep{molnar2022}. Since they correspond directly to input data, they align more naturally with human reasoning processes and are generally easier to interpret, including for medical professionals without specialized training in machine learning \citep{narayann2024protos}. Prototypes can serve as both \textit{local explanations} by showing cases similar to the predicted instance and as \textit{global explanations} by presenting representative examples from the data. This makes them a powerful tool for understanding both individual predictions and overall model reasoning.

Inspired by the paper~\cite{chen2019protpnet} on the \textit{prototypical part-based network} for image classification, where predictions are explained through interpretable patches rather than complete images, we explore how similar principles can be adapted for tabular medical data. Despite the success of \cite{chen2019protpnet} in other domains, prototype networks for tabular data remain underexplored, particularly in healthcare. This is notable because medical data often use discrete range language rather than raw features values. Discretized variables are easier to interpret because they correspond to clear, meaningful categories, such as age ranges, test result groups, or risk levels. These discrete features align better with clinical reasoning and allow for more transparent decision-making. Using these features, models can offer more intuitive explanations, helping physicians understand the predictions and relate them to real-world clinical scenarios.

To address this gap, we propose a prototype-based neural network, called \textit{Model for Explainable Diagnosis using Interpretable Concepts} (MEDIC), specifically designed for tabular medical data. Our approach introduces discrete prototypes, with the aim of improving interpretability while maintaining strong predictive performance. The goal of this study is to develop and evaluate this model in the context of medical records of patients, with a focus on producing faithful and physician-friendly explanations.

To ensure reproducibility, the code is publicly available in a GitHub repository\footnote{\url{https://github.com/jkarolczak/medic}}.

\section{Related work}

The work~\cite{christodoulou2019logisticregression} claims that deep neural networks~\cite{furnkranz2010encyclopedia} do not provide significant performance advantages over classical approaches such as random forest ~\cite{breiman2001rf} or gradient boosting (GB)~\cite{chen2016xgboost} for clinical prediction tasks utilizing tabular data, which may explain their limited adoption in the healthcare domain~\cite{christodoulou2019logisticregression}.

Despite machine learning models consistently demonstrating superior performance with tabular clinical data, these ensemble methods suffer from inherent opacity in their decision processes, creating a critical need for effective explanation frameworks that can provide healthcare professionals with transparent insights into model reasoning~\cite{adeniran2024xaihealthcare}.

The landscape of explainable AI approaches can be broadly categorized into two paradigms: \textit{feature attribution methods} and \textit{concept-based methods}~\cite{longo2024manifesto}. The first group was briefly discussed in Section~\ref{sec:introduction}. The second category includes \textit{prototype-based explanations} (also called example-based or instance-based explanations~\cite{bodria2023xai}), has shown particular promise in aligning with human cognitive processes, especially in domains where case-based reasoning is predominant. This is particularly true in the medical domain, where such methods have been shown to be effective in improving interpretability and trust \cite{bharati2024healthcarereview}.

Prototype-based explanations generally fall into two families. The post-hoc family identifies prototypes after model training, typically selecting representative instances from the training set. Notable algorithms include MMD-Critic \cite{kim2016mmd}, which employs maximum mean discrepancy to select prototypes and criticisms, and optimization-based approaches like A-PETE \cite{karolczak2024apete} and IKNN\_PSLFW \cite{zhang2022distantprotos}. Although straightforward to apply to tabular medical data, these methods often struggle with high-dimensional datasets containing many irrelevant features, which is a common characteristic in healthcare, where comprehensive diagnostic panels frequently generate information records containing redundant information~\cite{beam2017bigdata}. In this context, it is important to guide the decision maker’s attention toward the specific features of the prototype that the model considers relevant~\cite{karolczak2025conference}.

The second family, ante-hoc or intrinsic prototype methods, integrates prototype reasoning directly into the model architecture. Usually, these approaches represent prototypes not as complete instances but as parts or feature conjunctions that participate in decision making through mechanisms such as weighted voting. This direction gained significant attention following~\cite{li2018protonetwork}, where the approach was originally proposed for image classification.

ProtoPNet~\cite{chen2019protpnet} represents a breakthrough in this area, introducing a convolutional neural architecture where class predictions are based on similarities of the learned prototypical parts of images. The key innovation of ProtoPNet was enabling interpretability through visualization of prototypical image patches that the model "looks for" when making classifications. When classifying a new image, ProtoPNet identifies similar-looking patches in the input and compares them to its learned prototypes, with the similarity scores directly contributing to class predictions. This approach is particularly powerful for medical imaging, where specific visual patterns (such as tumors or lesions) are diagnostically significant. As documented in~\cite{santi2024protoreview} and demonstrated in applications like~\cite{singh2025medprot}, such models enhance transparency by highlighting medically relevant image regions and explicitly connecting them to learned prototypes that represent typical visual manifestations of conditions.

However, despite ProtoPNet's successful application across various image processing tasks~\cite{santi2024protoreview}, adapting this architecture for tabular data presents unique challenges. Medical tabular data lack the spatial structure of images that convolutional networks exploit, requiring fundamentally different approaches to identify meaningful "parts" or feature conjunctions. To date, a comparable architecture specifically designed for tabular medical records remains conspicuously absent from the literature.

\section{MEDIC: Model for Explainable Diagnosis using Interpretable Concepts}

In this section, for the first time, we propose the neural network \textit{MEDIC: Model for Explainable Diagnosis using Interpretable Concepts}. MEDIC is inspired by the prototypical parts paradigm proposed in~\cite{chen2019protpnet} and is designed to produce accurate and inherently interpretable predictions, which makes it particularly well suited for medical decision support, which usually requires human interpretation of proposed results. The model decomposes the decision support process into a small number of meaningful, human-understandable components: discretized input features describing the patient, interpretable feature subsets (parts) and class prototypes grounded in real data. These elements enable case-based reasoning and transparent justification of the proposed classification.

At a high level, the architecture follows an interpretable processing pipeline consisting of four key stages: (1) input discretization, which transforms continuous variables in the patient's description into symbolic bins; (2) part extraction, which identifies sparse and semantically coherent subsets of input features; (3) prototype comparison, where each extracted part of the patient's description is matched to learned prototypes, stored as embeddings representing features subsets of feature-value pairs from the training data; and (4) classification of the considered instance based on its similarity to prototypes. The complete MEDIC model is trained end-to-end to jointly learn all of these components in a supervised setting.

In clinical data, where features often come from heterogeneous sources (e.g. lab tests values, vital signs, diagnoses), such structured reasoning aligns well with domain expert expectations. Discretized bins can reflect clinically relevant ranges of diagnostic tests (e.g., abnormally high glucose), sparse parts mirror combinations that physicians would consider jointly (e.g., elevated CRP and fever), and prototypes anchor predictions in real cases that can be inspected post hoc.

We now describe the architecture in detail, starting with describing the interpretable discretization of continuous input features.

\subsection{Interpretable discretization of continuous input features}
\label{sec:discretization}

The decretization of continuous medical variables (e.g. age, lab tests values) into symbolic categories can aid interpretation and facilitate reasoning about patient features. In this work, our aim is to ultimately produce ranges of continuous features for symbolic interpretability. However, such a hard discretization is hardly optimizable in gradient-based neural network training.

To overcome this challenge, we introduce a fuzzy binning layer that enables a smooth, differentiable approximation of hard discretization during training. This allows gradients to flow through the discretization process and enables end-to-end optimization. After training, the soft representation can be replaced with a deterministic hard binning for better interpretability.

\paragraph{Fuzzy Binning}
\label{sec:fuzzy}
To allow interpretable discretization of continuous input features, we introduce a fuzzy binning layer that softly assigns each scalar feature value $x \in \mathbb{R}$ to a set of $K$ trainable bins. Each bin is characterized by a learnable center $\mu_k \in \mathbb{R}$ and a shared bandwidth parameter $\sigma > 0$. The soft membership of $x$ to the bin $k$ is defined using a Gaussian kernel:

\begin{equation}
    d_k(x) = \frac{(x - \mu_k)^2}{2\sigma^2}
\end{equation}

\begin{figure}
    \centering
    \begin{subfigure}[t]{0.45\textwidth}
        \centering
        \includegraphics{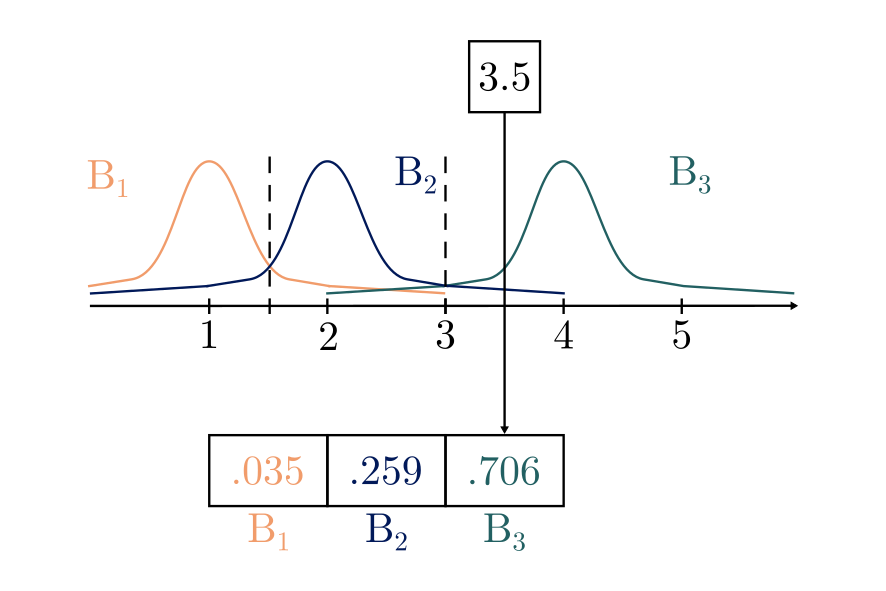}
        \caption{\textbf{Fuzzy binning}: the input value is softly assigned to each bin based on proximity to bin's center. The final encoding is a weighted combination, where the weights reflect similarity to the bin centers.}
        \label{fig:fuzzy-binning}
    \end{subfigure}
    ~
    \begin{subfigure}[t]{0.45\textwidth}
        \centering
        \includegraphics{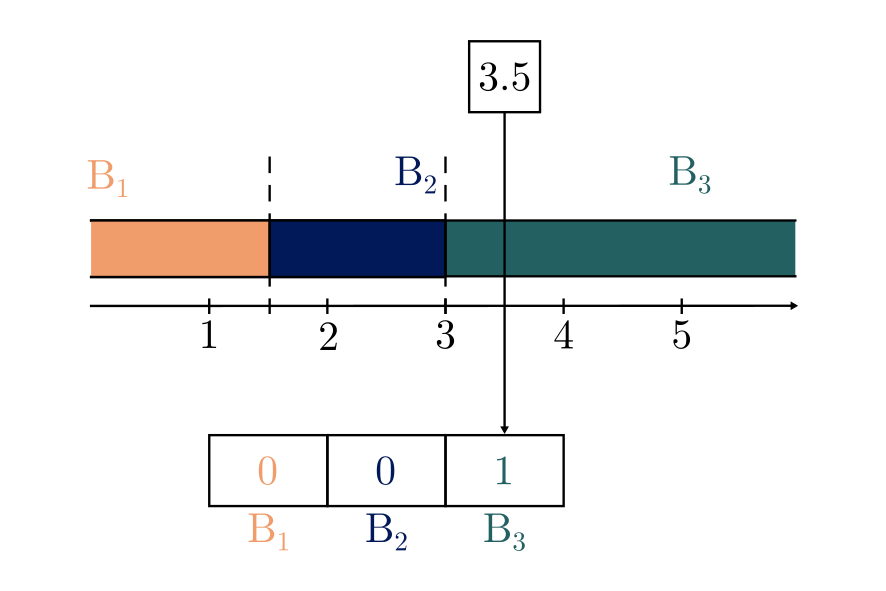}
         \caption{\textbf{Hard binning}: the input is deterministically assigned to a single bin with the nearest center. The encoding becomes a one-hot vector.}
        \label{fig:hard-binning}
    \end{subfigure}
    \caption{Comparison between fuzzy and hard binning for a scalar input value.}
    \label{fig:discretization}
\end{figure}

\begin{equation}
    \tilde{b}_k(x) = \frac{\exp(-d_k(x))}{\sum_{j=1}^{K} \exp(-d_j(x)) + \varepsilon}
\end{equation}
\noindent
where $\tilde{b}_k(x)$ denotes the normalized soft assignment and $\varepsilon$ is a small constant added for numerical stability. This results in a fuzzy, probabilistically weighted representation over bins, allowing each input to contribute partially to multiple bins (see Figure~\ref{fig:fuzzy-binning}).

The use of Gaussian kernels for fuzzy binning offers several advantages over direct distance-based assignment (e.g., $L^2$ norm). First, the smooth exponential decay naturally reflects uncertainty in proximity, which is especially relevant when feature values lie near bin boundaries. Second, the resulting softmax distribution is differentiable and normalized, facilitating gradient-based optimization in deep neural networks. 

Importantly, the bin centers $\mu_k$ and shared bandwidth $\sigma$ are optimized jointly with other model parameters during end-to-end training, allowing the discretization scheme to adapt to the data distribution.

\paragraph{Hard Binning}
\label{sec:hard}
After initial training of the network, the discretization is switched to the hard mode. In the hard setup, the input is assigned to a single bin via a non-differentiable $\arg\min$ operation over squared distances:

\begin{equation}
    \hat{b}(x) = \mathrm{one\_hot}\left( \arg\min_k \, (x - \mu_k)^2 \right)
\end{equation}

The resulting representation is a one-hot\footnotemark{} vector (Figure~\ref{fig:hard-binning}), which can be advantageous for symbolic interpretation and comparison of prototypes. However, it lacks gradient flow, making it unsuitable for end-to-end training.

In the hard binning regime, the input feature values are partitioned into contiguous intervals derived from the learned bin centers $\{\mu_k\}_{k=1}^K$. Specifically, each bin $k$ is associated with the interval
\begin{equation}
    I_k =
\begin{cases}
    \big( -\infty, \frac{\mu_1 + \mu_2}{2} \big) & \text{if } k = 1 \\
    \big[ \frac{\mu_{k-1} + \mu_k}{2}, \frac{\mu_k + \mu_{k+1}}{2} \big) & \text{if } 1 < k < K \\
    \big[ \frac{\mu_{K-1} + \mu_K}{2}, +\infty \big) & \text{if } k = K
\end{cases}
\end{equation}
\noindent
such that any scalar input \(x\) is discretized into the bin $k$ for which $x \in I_k$. This alternative representation facilitates interpretation by decision makers.

\subsection{MEDIC Architecture}
\label{sec:architecture}

\begin{figure}
    \centering
    \includegraphics{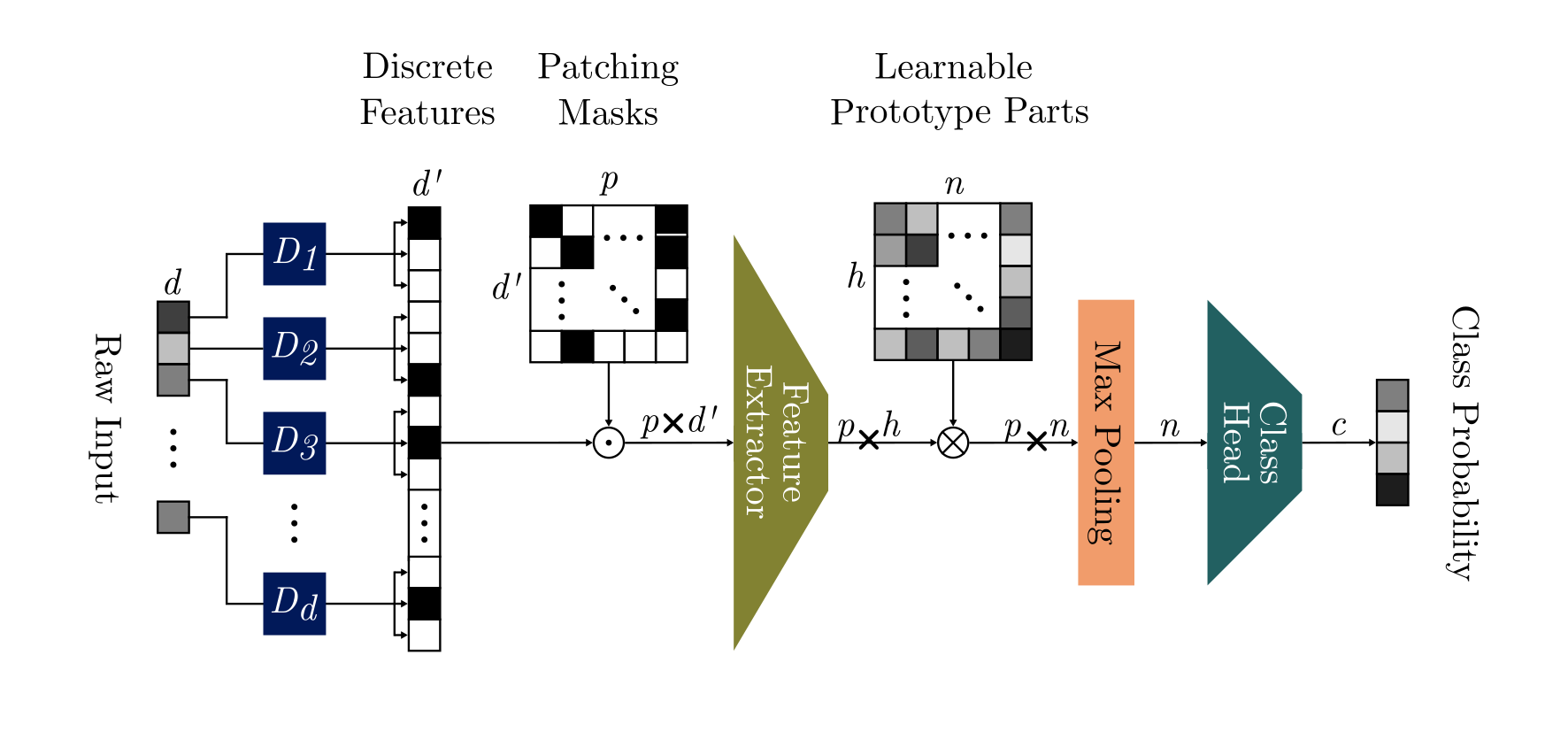}
    \caption{Overview of the Model for Explainable Diagnosis using Interpretable Concepts (MEDIC) architecture. Input features describing the patient are first discretized into binary features. These are projected using Hadamard product ($\odot$) into a set of $p$ interpretable input parts using sparse patching masks. Parts are encoded into an embedded space, compared to $n$ learnable prototypes using $L^2$ distance ($\otimes$), and the resulting distances are pooled and passed through a classification head to class assignment probabilities.}
    \label{fig:nn}
\end{figure}

MEDIC is a neural network inspired by the interpretable prototypical parts-based classification paradigm proposed in \cite{chen2019protpnet} and integrates symbolic input binning, feature extraction, prototype learning, and class prediction based on association with learned prototypes. The overview of the entire MEDIC architecture is shown in Figure~\ref{fig:nn}.

In the beginning, the raw input vector of features describing the patient is transformed by the network into a sparse high-dimensional binary representation. Each continuous feature is processed by a binning module introduced in Section~\ref{sec:discretization}. Meanwhile, categorical features undergo one-hot\footnotemark[\value{footnote}] encoding. All vectors coming from discretization are concatenated into a single $d'$ dimensional vector.

\footnotetext{A one-hot vector is a way of representing categories or intervals where only one entry is "on" (set to 1) and all others are "off" (set to 0), see~\ref{fig:hard-binning}. For example, if a lab result like albumin is split into three ranges (low, normal, high), only one of these will be marked as active. This makes it easy to interpret into which clinical range the value falls.}

Next, the binarized input is multiplied with a trainable set of $p$ patching masks, encoded as a matrix $M \in \mathbb{R}^{p \times d'}$. Each mask selects and linearly combines a sparse subset of binary features, effectively defining a part of the input instance (i.e. its description by features). Intuitively, each part can be seen as a meaningful combination of clinical indicators -- for example, \textit{high blood pressure in elderly patients} or \textit{elevated glucose and BMI}. This encourages the model to focus on patterns that are not only interpretable but also structured in a way that reflects domain knowledge.

Each of the $p$ part vectors is passed through a shared feature extractor module, implemented as a shallow multilayer perceptron with ReLU activations. This module transforms each sparse binary part into a dense embedding -- a compact vector of size $h$ that captures abstract and informative features. Embeddings are designed to preserve meaningful relationships in the data while reducing dimensionality, enabling the model to generalize across similar patterns. From an interpretability perspective, this step summarizes each clinically relevant pattern into a low-dimensional representation that retains the most diagnostically informative aspects.

To facilitate interpretable decision-making, the network maintains a set of $n$ learnable prototype vectors of size $h$. For each input part represented as embedding, the $L^2$ distance is computed for every prototype. This results in a $p \times n$ distance matrix, where each entry quantifies the dissimilarity between a specific part and a prototype. A max-pooling operation across parts selects the most relevant part for each prototype, yielding a vector of $n$ minimal distances, where each entry reflects the smallest distance between a given prototype and the most similar embedding representing a part of the input describing the patient. This enables comparison of each prototype to its best-matching part in the patient description.

To enable case-based reasoning, the network maintains a set of $n$ learnable prototype vectors, each of dimension $h$. Conceptually, each prototype represents a summary of a typical clinical condition or patient case learned from data. Each prototype is anchored in real patient data and corresponds to a representative example that lies near the center of a cluster of similar cases, making it reflective of common patterns observed across many patients. For every embedding corresponding to the part of patient description, the model computes the squared Euclidean ($L^2$) distance to each prototype, yielding a $p \times n$ distance matrix. Each row corresponds to one patient description part and each column to a prototype.

Then a maximum-pooling operation is applied across the rows of this matrix (that is, across parts), selecting for each prototype, the input part that has the smallest distance to that prototype, effectively identifying the most similar part. This produces a distance vector of length $n$, which summarizes how closely the input aligns with each of the learned prototypes.

Finally, this vector of distances is passed through a linear classification layer, producing a probability distribution over $c$ target classes. Since the classification decision is based directly on similarity to interpretable prototypes, each linked to specific input parts, the resulting predictions can be traced back and explained in terms of clinically meaningful comparisons to learned prototype parts.

\subsection{Three-Stage Training Procedure}

To ensure stable and interpretable network training, we adopt a three-stage training procedure that enables the model to learn hard bins -- intervals, and realistic prototype parts directly from the training data, all within a gradient-based optimization framework.

\paragraph{Stage 1: Initialization with Fuzzy Binning and Learnable Prototypes}
In the initial stage, the entire network is trained end-to-end with fuzzy binning and randomly initialized learnable prototypes. This setting ensures smooth gradient flow through the discretization modules, allowing the network to co-adapt binning thresholds and part extraction masks.

Fuzzy binning uses soft Gaussian kernels (Section~\ref{sec:fuzzy}, Figure~\ref{fig:fuzzy-binning}), which provide fuzzy assignments across bins. The patching masks and prototypes are trained jointly using classification loss, combined with auxiliary regularization terms: (1) L1 sparsity of patching masks, and (2) a diversity penalty to encourage spread among prototypes, which are further discussed in Section~\ref{sec:objective}.

\paragraph{Stage 2: Hard Binning and Mask Discretization}
Once convergence  in the training criterion is achieved, the discretization mode is switched to a hard mode by replacing the fuzzy binning with hard $\arg\min$ bin selection, freezing binning thresholds (Section~\ref{sec:hard}, Figure~\ref{fig:hard-binning}). Additionally, patching masks are binarized by thresholding to enforce strict binary groupings of input dimensions into parts.

This transition enables symbolic interpretability and highlights which specific input features are most relevant for each part. The rest of the network is fine-tuned using discretized inputs, preserving the interpretability of the parts.

\paragraph{Stage 3: Prototype Replacement with Real Parts}
Finally, the learned prototypes are replaced with embeddings derived from parts of actual patient records in the training data, ensuring that each prototype corresponds to a real and representative clinical case. For each prototype, the closest embedded input part is identified using the $L^2$ distance. These real parts are then copied into the prototype memory, replacing the synthetic prototypes. This step improves interpretability by anchoring each prototype to an actual example from the data.

This last step grounds the network's reasoning in actual data, allowing domain experts to inspect prototypical cases for each class. During this phase, the prototype embeddings are frozen, and only the classification head is fine-tuned to maintain stable performance, as accuracy would otherwise be expected to decline.

\subsection{Objective Function and Regularization}
\label{sec:objective}

The model is trained using cross-entropy loss, later denoted as $\mathcal{L}_{CE}$, as the standard objective for classification tasks. To improve interpretability and promote efficient structure, two regularization terms are added. The first is an $\ell_1$ sparsity penalty applied to the patching mask parameters:
\begin{equation}
    \mathcal{L}_{\text{sparsity}} = \lambda_{\text{sparsity}} \cdot \ell_1(M_{ij}) = \lambda_{\text{sparsity}} \cdot \frac{1}{p d'} \sum_{i=1}^{p} \sum_{j=1}^{d'} \left| M_{ij} \right|,
\end{equation}
where $M \in \mathbb{R}^{p \times d'}$ are the patching masks. This term encourages parts to rely on a minimal set of input features describing each patient. The $\ell_1$ penalty is chosen because, unlike $\ell_2$, it promotes exact zeros in patching masks, effectively turning off irrelevant input dimensions, and therefore leading to sparser and therefore more interpretable part-feature associations.

The second term encourages diversity among prototypes by penalizing redundancy in their representations:
\begin{equation}
    \mathcal{L}_{\text{diversity}} = - \lambda_{\text{diversity}} \cdot \frac{1}{N(N - 1)} \sum_{i \neq j} \left\| \mathbf{z}_i - \mathbf{z}_j \right\|_2,
\end{equation}
where $\mathbf{z}_i$ and $\mathbf{z}_j$ are prototype embeddings. This promotes coverage of distinct regions in the latent space. The full training objective function is the sum of three above-mentioned loss parts:
\begin{equation}
    \mathcal{L} = \mathcal{L}_{\text{CE}} + \mathcal{L}_{\text{sparsity}} + \mathcal{L}_{\text{diversity}}\,.
\end{equation}

\section{Experiments}

This section presents a comprehensive evaluation of our model, both from an interpretability and a predictive performance perspective. In the beginning, in Section~\ref{sec:predictive-perfromance} we assess the predictive accuracy of the method on three benchmark datasets, comparing its performance to selected baseline models. Subsequently, in Section~\ref{sec:case-study} we demonstrate how the learned prototypes may be applied in practice, through a case study grounded in a real-world medical dataset. This analysis highlights the interpretability of MEDIC and its ability to form clinically plausible representations. 

\subsection{Predictive performance}
\label{sec:predictive-perfromance}

\subsubsection{Experimental setup}
\label{sec:experimental-setup}

\paragraph{Data} 
To evaluate the proposed model, three publicly available medical datasets were selected: \textit{Cirrhosis}\footnote{\url{https://archive.ics.uci.edu/dataset/878/cirrhosis+patient+survival+prediction+dataset-1}}, \textit{Chronic Kidney Disease (CKD)}\footnote{\url{https://archive.ics.uci.edu/dataset/336/chronic+kidney+disease}}, and \textit{Diabetes}\footnote{\url{https://www.kaggle.com/datasets/mathchi/diabetes-data-set}}. These datasets were chosen due to their clinical relevance and inclusion of multiple laboratory measurements such as blood test results, namely:
\begin{itemize}
    \item Cirrhosis: bilirubin, cholesterol, albumin, copper, triglycerides, alkaline phosphatase (ALP), serum glutamic-oxaloacetic transaminase (SGOT), platelets, and prothrombin time;
    \item CKD: red blood cells, pus cells, pus cell clumps, blood glucose, blood urea, serum creatinine, sodium, potassium, hemoglobin, packed cell volume, white blood cell count, and red blood cell count;
    \item Diabetes: blood glucose and insulin.
\end{itemize}
Enumerated tests are well suited for discretization. These datasets also include additional numerical indicators, such as body mass index (BMI, in the Diabetes dataset), which further benefit from discretization by enhancing interpretability. Moreover, all three datasets exhibit a class imbalance: Cirrhosis contains 125, 19, and 168 instances for classes 0, 1, and 2 respectively; CKD consists of 115 and 43 instances for classes 0 and 1; and Diabetes includes 500 negative and 268 positive samples. These characteristics present a realistic benchmark for evaluating the model’s ability to process numerical medical features while addressing class imbalance, a common challenge in clinical predictive modeling~\cite{stefanowski2016challenges}.

\paragraph{Baselines}
\label{sec:baselines}
To evaluate the effectiveness of our prototype-based method, we compare it with a set of well-established baseline models commonly used in clinical machine learning tasks. Ensemble methods such as \textit{Random Forest (RF)}~\cite{breiman2001rf} and \textit{Gradient Boosting}, specifically the \textit{XGBoost (XGB)} implementation~\cite{chen2016xgboost}, serve as strong baselines due to their robustness, ability to capture non-linear feature interactions, and proven success in medical applications~\cite{vlachas2022rfmedicine}. We include a \textit{Decision Tree (DT)} model~\cite{furnkranz2010encyclopedia} as a reference for interpretability, since it represents models that are interpretable by design and is also a classifier that turned out to be sufficient to solve some problems~\cite{podgorelec2002decisiontree}.
Furthermore, we incorporate a simple feedforward neural network, also known as a \textit{Multi-Layer Perceptron (MLP)}~\cite{furnkranz2010encyclopedia}, to provide a baseline comparison within the class of neural models. The MLP consists of an input layer, one or more hidden layers with nonlinear activation functions, and an output layer with softmax activation for classification. The Decision Tree, Random Forest, and MLP, utilized implementations of the scikit-learn\footnote{\url{https://scikit-learn.org/}} Python package. The XGB implementation comes from the XGBoost\footnote{\url{https://xgboost.readthedocs.io/}} package.

\paragraph{Criterion}
\label{sec:criterion}
To compare the performance of different models, we use the geometric mean (g-mean) of sensitivity and specificity. This metric is especially useful in medical datasets that have class imbalance, where one outcome (e.g., disease presence) is much rarer than the other. Unlike accuracy, g-mean ensures that the model performs well on both positive and negative classes, making it a more balanced and clinically meaningful measure of performance \cite{brzezinski2018measures}.

\paragraph{Hyperparameter Optimization} 

Hyperparameter optimization (HPO) is essential to achieve strong and unbiased performance between models, particularly in settings that involve heterogeneous architectures. For all evaluated models, we used the \textit{Tree-structured Parzen Estimator Approach} (TPE) \cite{bergstra2011hpo} implemented in the Optuna framework\footnote{\url{https://optuna.org}} to perform black-box optimization of key hyperparameters. Each model was independently tuned using 100 optimization trials to maximize the g-mean metric~\cite{furnkranz2010encyclopedia}. To ensure a reliable estimation of predictive performance on unseen data, we adopt a 5-fold cross-validation framework.

For MEDIC, we recommend a structured tuning strategy informed by prior experience. Begin by setting $\lambda_{\text{sparsity}}$ to a relatively high value (e.g. $\approx 1.0$) and $\lambda_{\text{diversity}} = 0$, together with a large number of prototypes (e.g. 64). Gradually decrease $\lambda_{\text{sparsity}}$ until the number of activated features in the prototype parts stabilizes within a comprehensible range, ideally fewer than 5-7 features per prototype part. Once this is achieved, incrementally increase $\lambda_{\text{diversity}}$ to promote diversity inactivated parts, ensuring that the prototype part lengths remain consistent. After arriving at interpretable and stable prototype configurations, the remaining hyperparameters, including the number of prototypes (see Table~\ref{tab:hpo}) can be automatically tuned using a hyperparameter optimization algorithm such as TPE~\cite{bergstra2011hpo}.

Table~\ref{tab:hpo} summarizes the hyperparameters tuned for each model and the corresponding search spaces. For implementation-specific details, we refer the reader to the respective baseline packages cited in the \textbf{Baselines} paragraph above. 

\begin{table}[t]
    \centering
    \caption{Hyperparameters and the ranges of values allowed for selection during hyperparameter optimization.}
    \label{tab:hpo}
    \begin{tabular}{l | l l l}
        \toprule
        model & hyperparameter name & type & values  \\
        \midrule

        \multirow{6}{*}{DT} & ccp\_alpha & float & $[ 1e^{-5}, 0.1]$\\
        & criterion & categorical & \{ gini, entropy, log\_loss \} \\
        & max\_depth & integer & $[3, 30]$ \\
        & max\_features & categorical & \{ sqrt, log2, None \} \\
        & min\_samples\_leaf & integer & $[1, 20]$ \\
        & min\_samples\_split & integer & $[2, 20]$ \\       
        \hline
        
        \multirow{5}{*}{RF} 
        & max\_depth & integer & $[3, 30]$ \\
        & max\_features & categorical & \{ sqrt, log2, None \} \\
        & min\_samples\_leaf & integer & $[1, 20]$ \\
        & min\_samples\_split & integer & $[2, 20]$ \\
        & n\_estimators & integer & $[50, 300]$ \\
        \hline
        
        \multirow{9}{*}{XGB} & alpha & float & $[1e^{-3}, 10]$ \\
        & colsample\_bytree & float & $[0.5, 1]$ \\
        & gamma & float & $[0, 5]$ \\
        & lambda & float & $[1e^{-3}, 10]$ \\
        & learning\_rate & float & $[1e^{-3}, 0.3]$ \\
        & max\_depth & integer & $[3, 15]$ \\
        & min\_child\_weight & float & $[0.5, 10]$ \\
        & n\_estimators & integer & $[50, 500]$ \\
        & subsample & float & $[0.5, 1]$ \\
        \hline
        
        \multirow{5}{*}{MLP} & activation & categorical & \{ relu, tanh, logistic \} \\
        & alpha & float & $[ 1e^{-5}, 0.1]$ \\
        & hidden\_layer\_sizes & categorical & \{ 100,),\,(50, 50),\,(100, 50),\,(50, 25),\,(30, 30, 30) \} \\
        & learning\_rate\_init & float & $[ 1e^{-5}, 0.1]$ \\ 
        & max\_iter & integer & $[100,1000]$ \\
        \hline
        
        \multirow{4}{*}{MEDIC} & batch\_size & integer & \{ 16, 32, 64, 80, ..., 240, 256 \}  \\ 
        & hidden\_dim & int & $[2, 16]$ \\
        & learning\_rate & float & $[1e^{-5}, 0.1]$ \\
        & n\_prototypes & integer & \{ 4, 8, 12, 16, ..., 92, 96 \} \\
        \bottomrule
    \end{tabular}
\end{table}

\subsubsection{Results}
The performance of the model is summarized in Table~\ref{tab:gmean}, using the geometric mean (g-mean) metric in three datasets. MEDIC demonstrates competitive performance, achieving the best g-mean on the Cirrhosis and CKD datasets. In the diabetes data set, although XGB achieved the highest score, MEDIC followed closely, within less than a percentage point, indicating comparable effectiveness.

\begin{table}[t]
    \centering
    \caption{Classification performance (g-mean) across datasets. Bold values denote the best result for each dataset.}
    \label{tab:gmean}
    \begin{tabular}{l | c c c}
        \toprule
        & Cirrhosis & CKD & Diabetes \\
        \midrule
        DT        & 0.6564 & 0.9889 & 0.7327  \\
        RF        & 0.6742 & \textbf{1.0000} & 0.7432 \\
        XGB       & 0.6840 & 0.9889 & \textbf{0.7453} \\
        MLP       & 0.6765 & \textbf{1.0000} & 0.7381 \\
        MEDIC & \textbf{0.6889} & \textbf{1.0000} & 0.7367  \\
        \bottomrule
    \end{tabular}
\end{table}

Table~\ref{tab:nprototypes} shows the maximum allowed number of prototypes defined as a model setting (hyperparameter) and the number of unique prototype parts actually discovered by the MEDIC model during training. The results suggest that the model can self-regularize by reusing the same prototype multiple times when no additional meaningful feature-value sets (prototype parts) can be identified. This indicates that MEDIC avoids overfitting by focusing only on truly informative patterns, even when more prototypes are allowed.

\begin{table}[t]
    \centering
    \caption{Comparison between the maximum number of parts allowed for the model (hyperparameter) and the number of unique prototype parts actually discovered by MEDIC for the optimal configuration found during hyperparameter optimization.}
    \label{tab:nprototypes}
    \begin{tabular}{l | c c c}
        \toprule
        & Cirrhosis & CKD & Diabetes \\
        \midrule
        Max. allowed & 52 & 40 & 44 \\
        Unique found & 14 & 10 & 8 \\
        \bottomrule
    \end{tabular}
\end{table}

\subsection{Studying MEDIC in action}
\label{sec:case-study}

To show that MEDIC is interpretable and present how prototype parts learned by MEDIC look like, we conducted a qualitative analysis using a case study for a single dataset -- Cirrhosis. Our aim is to illustrate how the MEDIC reasoning process works by comparing individual patient cases to representative clinical patterns (prototypes) previously learned from the data.

Table~\ref{tab:norms} shows the discretized feature intervals identified by the network. These intervals represent meaningful partitions of the input space and often align with known clinical thresholds. For reference, we compare them with the standard clinical intervals provided by the American College of Clinical Pharmacy\footnote{\url{https://www.accp.com/docs/sap/Lab_Values_Table_PSAP.pdf}}.

For example, the learned limit between intervals 1 and 2 for albumin is 3.7 g/dL, which closely matches the clinical lower limit of 3.5 g/dL. Similarly, the learned limits for the prothrombin time (10.52-10.93 seconds) are well within the reference range of 10–13 seconds. Triglycerides also have a limit near 137 mg / dL, close to the reference value of <150 mg/dL.

Earlier in this work, we justified using three bins per feature to intuitively capture low–normal–high ranges. However, for certain features in specific disease contexts, deviations in one single direction may be clinically significant. Interestingly, in this experiment the observations suggest the network to exhibit the ability to self-organize and adjust these bins accordingly. For example, for copper, the first interval is effectively disabled by learning a negative upper bound (–8.98), which is not physiologically plausible, thus disregarding it. Likewise, for platelets and albumin, the network forms exceptionally narrow middle intervals, suggesting that small changes within this range may be critical for the classification.

Although Table~\ref{tab:norms} shows intervals ranging from $-\infty$ to $\infty$ for technical completeness, in practical applications these can be translated into clinically relevant and bounded intervals. For example, lower limits can be set to zero and upper limits can be capped according to known physiological limits, without affecting the model’s learning performance. This translation can support physicians in interpreting the model’s behavior more easily.
\begin{table}[t]
    \centering
    \caption{Discretized feature intervals learned by the MEDIC network on the Cirrhosis dataset.}
    \label{tab:norms}
    \begin{tabular}{l | l l l}
        \toprule
        Feature & Interval 1 & Interval 2 & Interval 3 \\
        \midrule
        Albumin                 & $(-\infty, 3.70)$ & $[3.70, 3.82)$ & $[3.82, \infty)$ \\
        ALP                     & $(-\infty, 3668)$ & $[3668, 4113)$ & $[4113, \infty)$ \\
        Bilirubin               & $(-\infty, 0.79)$ & $[0.79, 3.43)$ & $[3.43, \infty)$ \\
        Cholesterol             & $(-\infty, 345)$ & $[345, 668)$ & $[668, \infty)$ \\
        Copper                  & $(-\infty, -8.98)$ & $[-8.98, 103.76)$ & $[103.76, \infty)$ \\
        N\_days                 & $(-\infty, 2152)$ & $[2152, 2343)$ & $[2343, \infty)$ \\
        Platelets               & $(-\infty, 271)$ & $[271, 291)$ & $[291, \infty)$ \\
        Prothrombin             & $(-\infty, 10.52)$ & $[10.52, 10.93)$ & $[10.93, \infty)$ \\
        SGOT                    & $(-\infty, 80)$ & $[80, 144)$ & $[144, \infty)$ \\
        Tryglicerides           & $(-\infty, 137)$ & $[137, 172)$ & $[172, \infty)$ \\
        \bottomrule
    \end{tabular}
\end{table}

Subsequently, MEDIC identified several prototype parts, specific combinations of clinical features and value ranges, that it considers informative to predict patient outcomes related to cirrhosis. These prototype parts are presented in Table~\ref{tab:prototypes}.

\begin{table}[t]
\centering
\caption{Prototype parts identified by the MEDIC for cirrhosis dataset.}
\label{tab:prototypes}
\begin{tabular}{c p{12cm}}
    \toprule
    & Prototype part \\
    \midrule
    1 & 
        \begin{minipage}[t]{12cm}
           Bilirubin in [0.79, 3.43) $\land$  Hepatomegaly = 0 $\land$ Spiders = 0 
        \end{minipage} \\
    2 & 
        \begin{minipage}[t]{12cm}
            Albumin $\in$ [3.82, $\infty$) $\land$ Bilirubin $\in$ [0.79, 3.43) $\land$ Hepatomegaly = 0 $\land$ Spiders = 0  
        \end{minipage} \\
    3 & 
        \begin{minipage}[t]{12cm}
            Cholesterol $\in$ [667, $\infty$) $\land$ Copper $\in$ [103.76, $\infty$) $\land$ Hepatomegaly = 0 $\land$ Spiders = 0
        \end{minipage} \\
    4 & 
        \begin{minipage}[t]{12cm}
            Bilirubin $\in$ ($-\infty$, 0.79) $\land$ Cholesterol $\in$ ($-\infty$, 345) $\land$  Hepatomegaly = 1 $\land$ Spiders = 0 
        \end{minipage} \\
    5 & 
        \begin{minipage}[t]{12cm}
            Albumin $\in$ [3.70, 3.82) $\land$ Bilirubin $\in$ [0.79, 3.43) $\land$ Cholesterol $\in$ ($-\infty$, 345)  $\land$ Hepatomegaly = 0 $\land$ Spiders = 0 
        \end{minipage} \\
    6 & 
        \begin{minipage}[t]{12cm}
            Bilirubin $\in$ [0.79, 3.43) $\land$ Cholesterol $\in$ ($-\infty$, 345) $\land$ Hepatomegaly = 0 $\land$  Platelets $\in$ ($-\infty$, 271) $\land$ Spiders = 0
        \end{minipage} \\
    7 & 
        \begin{minipage}[t]{12cm}
            Bilirubin $\in$ ($-\infty$, 0.79) $\land$ Cholesterol $\in$ ($-\infty$, 345) $\land$ Hepatomegaly = 0 $\land$ Platelets $\in$ ($-\infty$, 271) $\land$ Spiders = 0
        \end{minipage} \\
    8 & 
        \begin{minipage}[t]{12cm}
            Albumin $\in$ [3.70, 3.82) $\land$ Bilirubin $\in$ [0.79, 3.43) $\land$ Cholesterol $\in$ ($-\infty$, 345) $\land$ Hepatomegaly = 0 $\land$ Spiders = 0  
        \end{minipage} \\
    9 & 
        \begin{minipage}[t]{12cm}
            Albumin $\in$ [3.82, $\infty$) $\land$ Bilirubin $\in$ [0.79, 3.43) $\land$ Cholesterol $\in$ ($-\infty$, 345) $\land$ Hepatomegaly = 1 $\land$ Spiders = 0 
        \end{minipage} \\
    10 & 
        \begin{minipage}[t]{12cm}
            ALP $\in$ ($-\infty$, 3668) $\land$ Bilirubin $\in$ [0.79, 3.43) $\land$ Hepatomegaly = 0 $\land$ N\_Days $\in$ [2343, $\infty$) $\land$ SGOT $\in$ [80, 144) $\land$ Tryglicerides $\in$ ($-\infty$, 137)
        \end{minipage} \\
    11 & 
        \begin{minipage}[t]{12cm}
            ALP $\in$ ($-\infty$, 366) $\land$ Bilirubin $\in$ [0.79, 3.43) $\land$ Drug = 1 $\land$ Hepatomegaly = 0 $\land$ N\_Days $\in$ ($-\infty$, 2152) $\land$ SGOT $\in$ [80, 144) $\land$ Tryglicerides $\in$ [172, $\infty$)
        \end{minipage} \\
    12 & 
        \begin{minipage}[t]{12cm}
            Albumin $\in$ [3.82, $\infty$) $\land$ ALP $\in$ ($-\infty$, 3668) $\land$ Drug = 1 $\land$ Hepatomegaly = 0 $\land$ N\_Days $\in$ [2343, $\infty$) $\land$ SGOT $\in$ ($-\infty$, 80) $\land$ Tryglicerides $\in$ [172, $\infty$)
        \end{minipage} \\
    13 & 
        \begin{minipage}[t]{12cm}
            ALP $\in$ ($-\infty$, 3668) $\land$ Bilirubin $\in$ [0.79, 3.43) $\land$ Drug = 1 $\land$ Hepatomegaly = 0 $\land$ N\_Days $\in$ ($-\infty$, 2152) $\land$ Prothrombin $\in$ ($-\infty$, 10.52) $\land$ SGOT $\in$ [80, 144) $\land$ Tryglicerides $\in$ [172, $\infty$)
        \end{minipage} \\
    14 & 
        \begin{minipage}[t]{12cm}
            Albumin $\in$ [3.82, $\infty$) $\land$ ALP $\in$ ($-\infty$, 3668) $\land$ Bilirubin $\in$ [0.79, 3.43) $\land$ Drug = 1 $\land$ Hepatomegaly = 0 $\land$ N\_Days $\in$ [2343, $\infty$) $\land$ Prothrombin $\in$ ($-\infty$, 10.52) $\land$ SGOT $\in$ [80, 144) $\land$ Tryglicerides $\in$ ($-\infty$, 137) 
        \end{minipage} \\
    \bottomrule
    \end{tabular}
\end{table}

Next, we investigate how a specific patient case (shown in Table~\ref{tab:explained-instance}, classified as 0 -- death) is internally processed and classified by MEDIC through its similarity to the nearest learned prototype parts. Each prototype part consists of a sparse conjunction of conditions over discretized or binary features, typically involving only a small number of dimensions. For example, a prototype part may specify conditions such as: Bilirubin level within [0.79, 3.43) mg/dL, absence of hepatomegaly (Hepatomegaly = 0), and drug usage indicated (Drug = 1). These concise feature subsets capture clinically meaningful patterns that contribute to the model’s decisions. MEDIC’s classification is driven by the similarity between the patient's description and the prototype parts, which are easily accessible and can be examined by the user, thereby offering transparent insight into the model’s reasoning process.

The list of prototypes with the highest similarity to this example demonstrates how the model constructs its reasoning by combining interpretable substructures. Many of these substructures align with known clinical heuristics or highlight relevant feature interactions. For example, bilirubin, ALP (alkaline phosphatase), and N\_Days (duration since patient registration) appear frequently in the most similar prototypes, highlighting their importance as clinical indicators influencing classification.

\begin{table}[t]
    \centering
    \caption{Feature values of the described patient case. According to MEDIC, outcome for this patient was 0 -- death.}
    \label{tab:explained-instance}
    \begin{tabular}{l | c c c c c c c c c c c c c c c c}
        \toprule
        Feature & \rotatebox{90}{Albumin} & \rotatebox{90}{ALP} & \rotatebox{90}{Ascites} & \rotatebox{90}{Bilirubin} & \rotatebox{90}{Cholesterol} & \rotatebox{90}{Copper} & \rotatebox{90}{Drug} & \rotatebox{90}{Edema} &  \rotatebox{90}{Hepatomegaly} & \rotatebox{90}{N\_Days} & \rotatebox{90}{Platelets} & \rotatebox{90}{Prothrombin} & \rotatebox{90}{SGOT} & \rotatebox{90}{Spiders} & \rotatebox{90}{Tryglicerides} \\
        \midrule
        Value & 2.27 & 728 & 1 & 0.8 & 370 & 42 & 1 & 1 & 1 & 1217 & 156 & 11 & 71 & 0 & 125 \\
        \bottomrule
    \end{tabular}
\end{table}

These prototypes highlight clinically relevant signals, such as low bilirubin, no hepatomegaly, and shorter hospital stays (N\_Days), as contributors to the classification. Furthermore, the inclusion of interaction patterns, such as elevated triglycerides in the context of certain ranges of liver enzymes, reflects how the network captures more nuanced decision logic than simple thresholding.

\begin{enumerate}
    \item \textbf{Similarity}: 0.864 \\
    \textbf{Prototype:} \underline{N\_Days $\in$ ($-\infty$, 2152)} $\land$ \underline{Drug = 1} $\land$ \underline{Hepatomegaly = 0} $\land$ \underline{Bilirubin $\in$ [0.79, 3.43)} $\land$ \underline{ALP $\in$ ($-\infty$, 366)} $\land$ SGOT $\in$ [80, 144) $\land$ Tryglicerides $\in$ [172, $\infty$)
    \item \textbf{Similarity}: 0.846 \\
    \textbf{Prototype:} Hepatomegaly = 0 $\land$ \underline{Spiders = 0} $\land$ Cholesterol $\in$ [667, $\infty$) $\land$ Copper $\in$ [103.76, $\infty$)
    \item \textbf{Similarity:} 0.834 \\
    \textbf{Prototype:} N\_Days $\in$ [2343, $\infty$) $\land$ Hepatomegaly = 0 $\land$ \underline{Bilirubin $\in$ [0.79, 3.43)} $\land$ \underline{ALP $\in$ ($-\infty$, 3668)} $\land$ SGOT $\in$ [80, 144) $\land$ \underline{Tryglicerides $\in$ ($-\infty$, 137)}
    \item \textbf{Similarity:} 0.824 \\
    \textbf{Prototype:} N\_Days $\in$ [2343, $\infty$) $\land$ \underline{Drug = 1 }$\land$ Hepatomegaly = 0 $\land$ \underline{Bilirubin $\in$ [0.79, 3.43)} $\land$ Albumin $\in$ [3.82, $\infty$) $\land$ \underline{ALP $\in$ ($-\infty$, 3668)} $\land$ SGOT $\in$ [80, 144) $\land$ \underline{Tryglicerides $\in$ ($-\infty$, 137)} $\land$ Prothrombin $\in$ ($-\infty$, 10.52)
    \item \textbf{Similarity:} 0.798 \\
    \textbf{Prototype:} \underline{N\_Days $\in$ ($-\infty$, 2152)} $\land$ \underline{Drug = 1} $\land$ Hepatomegaly = 0 $\land$ \underline{Bilirubin $\in$ [0.79, 3.43)} $\land$ \underline{ALP $\in$ ($-\infty$, 3668)} $\land$ SGOT $\in$ 80, 144) $\land$ Tryglicerides $\in$ [172, $\infty$) $\land$ Prothrombin $\in$ ($-\infty$, 10.52)
\end{enumerate}

\section{Conclusions}

This work introduced MEDIC, a novel prototype parts-based neural network architecture that transforms the approach to interpretability in machine learning for medical tabular data. Unlike conventional post-hoc explanation methods that retrospectively justify black-box decisions, MEDIC represents a paradigm shift toward inherently interpretable models that mimic clinical reasoning patterns. The core innovation lies in our three-component architecture: (1) differentiable discretization that aligns with medical thresholds, (2) sparse patching masks that identify clinically meaningful feature combinations, and (3) prototype-based reasoning that grounds predictions in case-based comparisons, all unified within an end-to-end trainable framework.

Evaluation across three clinical datasets demonstrated that MEDIC achieves competitive and sometimes superior predictive performance compared to established methods while providing transparent decision processes. In particular, the model autonomously discovered discretization thresholds that closely align with clinically recognized reference ranges, as evidenced in our cirrhosis case study where albumin and prothrombin intervals closely matched established medical guidelines. Furthermore, the prototype parts learned by the model reflected combinations of features that correspond to recognizable diagnostic patterns, suggesting that MEDIC captures meaningful representations of clinical knowledge.

The implications of this work extend beyond technical innovation only. By bridging the gap between accuracy and interpretability, MEDIC addresses a critical barrier to AI adoption in healthcare, the lack of interpretability, which undermines the trust of clinicians and regulatory acceptance. Our approach supports collaborative human-AI decision making where the model's reasoning can be verified, critiqued, and integrated with clinical expertise.

Several promising directions emerge for future research. First, incorporating domain-specific prior knowledge into prototype learning could further align representations with medical understanding. Second, dynamic prototype adaptation would enable the model to update representations in response to symptom drift caused by evolving diseases or treatments~\cite{tran2022covid}. Finally, rigorous user studies with physicians would provide insights into how this approach affects clinical decision-making and how explanations could be optimized for maximum utility.

\begin{acknowledgments}
    This research was funded in part by National Science Centre, Poland OPUS grant no. 2023/51/B/ST6/00545 and in part by PUT SBAD 0311/SBAD/0752 grant.
\end{acknowledgments}

\bibliography{bibliography}



\end{document}